\documentclass[10pt,twocolumn,letterpaper]{article}

\usepackage[algorithms]{wacv} 

\usepackage{amsthm} 
\usepackage{thmtools} 
\usepackage{thm-restate} 

\usepackage{multirow}
\usepackage{graphicx}
\usepackage{arydshln}
\usepackage{diagbox}
\usepackage{booktabs}
\usepackage{siunitx}

\usepackage[table]{xcolor} 

\usepackage{bbm} 

\definecolor{wacvblue}{rgb}{0.21,0.49,0.74}
\usepackage[pagebackref,breaklinks,colorlinks,allcolors=wacvblue]{hyperref}

\def\wacvPaperID{2554} 
\def\confName{WACV}
\def\confYear{2026}

\title{Reliable and Reproducible Demographic Inference for Fairness in Face Analysis }

\usepackage{orcidlink}

\author{Alexandre Fournier-Montgieux$^{*1}$\orcidlink{0009-0002-7744-3179}\\
{\tt\small alexandre.fourniermontgieux@cea.fr}
\and 
Hervé Le Borgne$^1$\orcidlink{0000-0003-0520-8436}\\
{\tt\small  herve.le-borgne@cea.fr}
\and
Adrian Popescu$^1$\orcidlink{0000-0002-8099-824X}\\
{\tt\small adrian.popescu@cea.fr}
\and
Bertrand Luvison$^1$\orcidlink{0000-0003-2475-3712}\\
{\tt\small bertrand.luvison@cea.fr}
\and 
{\small $^1$Université Paris-Saclay, CEA, LIST,F-91120, Palaiseau, France}
}

\begin{document}
\maketitle

Fairness evaluation in face analysis systems (FAS) typically depends on automatic demographic attribute inference (DAI), which itself relies on predefined demographic segmentation. However, the validity of fairness auditing hinges on the reliability of the DAI process. We begin by providing a theoretical motivation for this dependency, showing that improved DAI reliability leads to less biased and lower-variance estimates of FAS fairness. 
To address this, we propose a fully reproducible DAI pipeline that replaces conventional end-to-end training with a modular transfer learning approach. Our design integrates pretrained face recognition encoders with non-linear classification heads. We audit this pipeline across three dimensions: accuracy, fairness, and a newly introduced notion of robustness, defined via intra-identity consistency. The proposed robustness metric is applicable to any demographic segmentation scheme.
We benchmark the pipeline on gender and ethnicity inference across multiple datasets and training setups. Our results show that the proposed method outperforms strong baselines, particularly on ethnicity, which is the more challenging attribute.
To promote transparency and reproducibility, we will publicly release the training dataset metadata, full codebase, pretrained models, and evaluation toolkit. This work contributes a reliable foundation for demographic inference in fairness auditing.

\section{Introduction}
\label{sec:intro}

The accuracy of automatic face analysis systems has significantly progressed over the past decade, driven by increasingly sophisticated deep learning architectures~\cite{deng2019arcface,kim2022adaface} and the availability of large-scale training datasets~\cite{zhu2023webface260m}.
FAS now powers critical applications, from smartphone authentication to national security infrastructure~\cite{Apple_FaceID_Whitepaper_2017, GAO_21_526_2021,Stodder_Warrick_Atlantic_2022}.
Given this societal impact, it must be not only accurate but also fair to ensure responsible deployment and public trust~\cite{raji2020saving}.
Numerous studies have shown that FAS performance can vary significantly across demographic groups~\cite{Buolamwini2018GenderShades,fournier2025fairer,Grother2019FRVTPart3,gwilliam2021rethinking,kotwal2025review,sarridis2023towards}.
These disparities are often attributed to demographic imbalance in training data~\cite{gwilliam2021rethinking,Wu_2023_BMVC}, limited data diversity~\cite{fournier2025fairer}, visual variability~\cite{terhorst2021comprehensive}, or suboptimal training procedures~\cite{krishnapriya2020issues}.
While these studies have advanced our understanding of algorithmic bias, they largely overlook a foundational assumption: the reliability of the demographic attribute inference (DAI) process itself.

\begin{figure}[t]
    \centering
    \includegraphics[width=\linewidth]{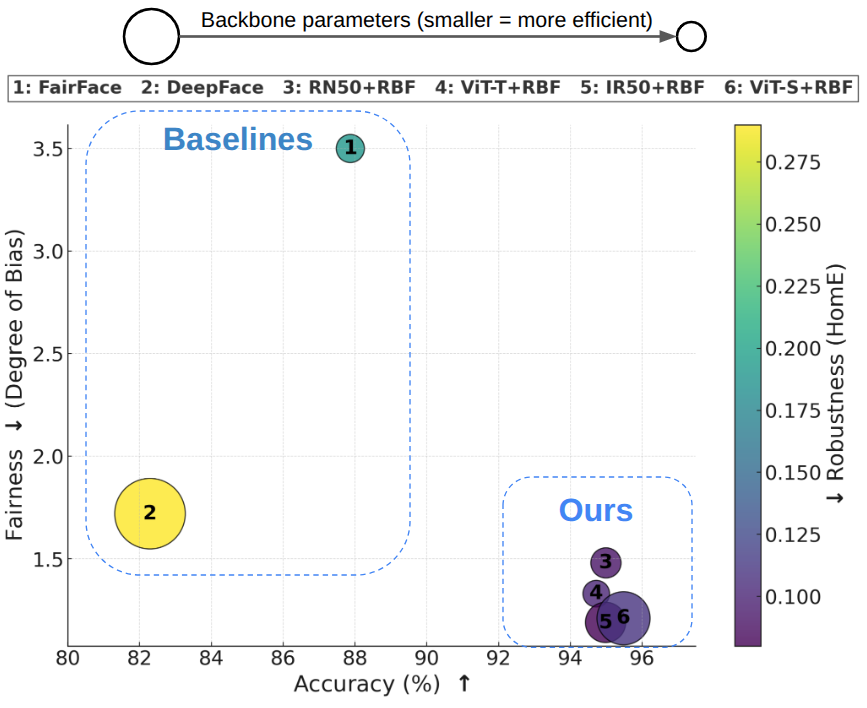}
    \caption{Ethnicity inference results (4 classes) audited using accuracy, fairness, and robustness dimensions on the BFW dataset~\cite{robinson2020face} for baselines (FairFace~\cite{Karkkainen2021FairFace} and DeepFace~\cite{serengil2024benchmark}) and several instances of the proposed approach. Higher values are better for accuracy. Lower values are better for fairness. Darker is better for robustness. The size of the circles is proportional to the number of parameters of the backbone DNN used, with smaller circles indicating improved inference efficiency. An ideal DAI system lies at the bottom-right corner with a collapsed circle.}
    \label{fig_teaser}
\end{figure}

In practice, attribute inference typically relies on off-the-shelf models~\cite{kotwal2025review}, with FairFace~\cite{Karkkainen2021FairFace} being the most widely used.
Yet, as shown in \autoref{fig_teaser}, FairFace offers suboptimal performance and compromises the fairness evaluation of downstream systems.
Compounding the issue, FairFace is difficult to audit: its training code is unavailable, and several reproduction attempts have failed~\footnote{\url{https://github.com/joojs/fairface/issues}}.

We begin by offering a theoretical justification for the need for reliable DAI. We then introduce a fully transparent and reproducible pipeline that jointly audits DAI performance across three dimensions: accuracy, fairness, and robustness.
Maximizing performance along all three is essential: accuracy underpins valid fairness audits; fairness ensures equitable error distribution across groups; robustness, an underexplored property, promotes intra-identity consistency.
We start from the following robustness-related hypothesis: 
\textit{Given $N$ images of the same identity, an ideal demographic classifier should predict the same segment for all samples.} 
We formalize this intuition using an entropy-based robustness metric that generalizes across demographic taxonomies and assumes only intra-identity consistency.

Unlike prior approaches such as~\cite{Karkkainen2021FairFace}, which rely on end-to-end training, we adopt a modular design: we pretrain or reuse face recognition backbones and train only the classification heads.
Using publicly available data, we conduct extensive experiments for ethnicity and gender inference, varying backbones, dataset configurations, and classification heads.
The ethnicity inference results in \autoref{fig_teaser} show significant accuracy, fairness, and robustness improvements for our approach compared with FairFace~\cite{Karkkainen2021FairFace} and DeepFace~\cite{serengil2024benchmark} with comparable backbones.
We observe further gains when using larger backbones. 

This work advances demographic attribute inference by making it more reliable and reproducible—two qualities that are critical for trustworthy fairness evaluations in face analysis.
To facilitate adoption and future research, we will release the full codebase, data processing scripts, evaluation protocols, DAI models, and plug-and-play inference tools.

\section{Related Work}
\label{sec:related}

Fairness evaluation ~\cite{Buolamwini2018GenderShades,fournier2025fairer,Grother2019FRVTPart3,gwilliam2021rethinking,raji2020about_face,sarridis2023towards} is needed to comprehensively audit AI-driven face analysis systems deployed in practical applications, including border control~\cite{hidayat2024face}, consumer authentication~\cite{teoh2021face}, public safety~\cite{miethe2025facial}, face-based payment~\cite{rija2022payment}.
These studies highlight significant performance variation across demographic segments, due to imbalance in face recognition training datasets~\cite{gwilliam2021rethinking, Wu_2023_BMVC}, insufficient data diversity~\cite{fournier2025fairer}, or biases in the architectures or training strategies~\cite{dooley2023rethinking,nagpal2019deep,rezgui2024controllable}. 
While understanding and mitigating biases is important, little effort is devoted to the reliability of the demographic attribute inference that enables their conclusions.
Below, we analyze the main components of existing inference pipelines, as well as other relevant works. 

The demographic segmentation scheme represents the basis for any demographic classifier. 
The dominant approach~\cite{Buolamwini2018GenderShades,Karkkainen2021FairFace,Wang2019RFW,wang2021meta} defines a small set of predefined ethnicity categories.
Past works proposed alternatives to the ethnicity-based categorization used in FairFace, notably based on skin color~\cite{chardon1991skin}.
The DiF dataset~\cite{merler2019diversity} implemented a facial coding scheme consisting of skin color and six other attributes describing the face shape. 
More recently~\cite{Thong2023BeyondSkinTone,lin2025ai_face} modeled skin appearance by adding the hue angle to the tone measured using the Fitzpatrick scale.
These segmentations aim to encode medical and/or anthropological attributes in face coding schemes, rather than predefining ethnic categories. 
This modeling choice makes them appealing alternatives, but existing studies do not test their robustness. 
In addition, their semantic interpretation is less direct compared to that of predefined categories.
We note that all demographic segmentation schemes are criticizable because they categorize the vast diversity of human faces into a limited set of categories.
The sensitive nature of face data amplifies the difficulty~\cite{kotwal2025review}. 
However, attribute segmentation is necessary to facilitate downstream evaluation. 
We argue that we should select a particular scheme based on a comprehensive analysis of its accuracy, robustness, and fairness. 

Training datasets are a core component of demographic classification.
In supervised settings, their annotation follows a predefined segmentation. 
Representative examples include the RFW~\cite{Wang2019RFW}, BFW ~\cite{robinson2020face}, FairFace~\cite{Karkkainen2021FairFace}, BUPT~\cite{wang2021meta}. 
Their common characteristics include the manual curation of demographic annotations and the emphasis on segment balancing. 
While balancing usually increases fairness, multiple studies~\cite{albiero2020analysis,fournier2025fairer,krishnapriya2020issues,sarridis2023towards} show that it only partially addresses DAI unfairness.
The authors of~\cite{terhorst2021comprehensive} investigate the contribution of non-demographic factors on face recognition biases. 
Different works adapted the training process to improve fairness: SensitiveLoss~\cite{serna2022sensitive}, a discrimination-aware loss, penalizes demographic disparities, GAC~\cite{gong2021mitigating} adapts convolution kernels and attention mechanisms to faces based on their demographic attributes, and DebFace~\cite{gong2020jointly} uses adversarial learning to reduce face recognition and DAI biases jointly, while~\cite{alvi2018turning} tackles attribute combinations during training. 
Interestingly, despite strong efforts devoted to face generation~\cite{Kim2023DCFace,lin2025ai_face}, fairness issues subsist and are even amplified compared with real data~\cite{fournier2025fairer}.
The authors of~\cite{Wu_2023_BMVC} examine the dataset structure (number of identities and of images per identity), but also factors DAI fairness, such as head pose, brightness, and image quality.
These studies inform our experiments, and we test different training dataset structures and sizes. 


FairFace~\cite{Karkkainen2021FairFace}, the work closest to ours, is a widely used demographic classification tool~\cite{d2024openbias,deandres2024good,fournier2025fairer,lee2024vhelm,melzi2024frcsyn,sarridis2023towards} that jointly infers ethnicity, gender, and age.
The authors demonstrate that training on their balanced dataset yields more equitable performance across demographic groups compared to models trained on imbalanced alternatives. 
One important limitation is that the method achieves 88\% 4-ethnicity classification accuracy on the RFW dataset, which affects the relevance of fairness auditing based on it. 
Beyond performance limitations, FairFace suffers from a critical reproducibility gap.
The authors have not released training code or sufficient implementation details to enable independent replication, despite numerous attempts~\footnote {\url{https://github.com/joojs/fairface/issues}}, including ours. 
This opacity undermines the scientific reproducibility that is essential for bias auditing, leaving practitioners unable to verify claimed performance or adapt the methodology to new contexts.
More recently, the DeepFace~\cite{serengil2024benchmark} framework includes DAI capabilities.
The demographic classifiers training merges the FairFace~\cite{Karkkainen2021FairFace} and UTKFace~\cite{zhang2017age} datasets to improve global accuracy.
We use FairFace and DeepFace as baselines in our experiments. 
\section{Demographic attribute inference}\label{sec:method}
\begin{figure*}[t]
    \centering
    \includegraphics[width=\linewidth]{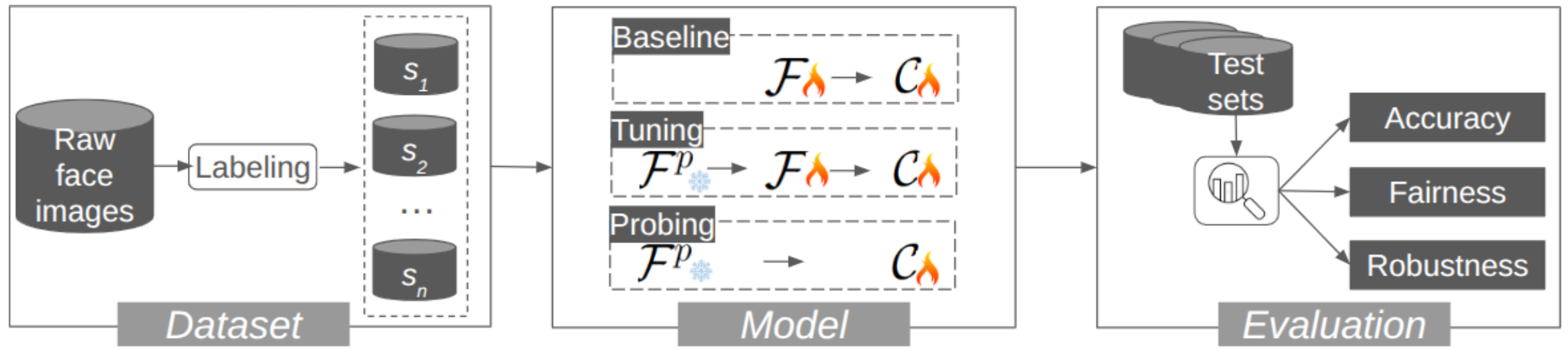}
    \caption{Overview of pipelines designed for training and evaluating demographic attribute inference. The raw data labels correspond to a demographic split schema comprising $n$ segments. The model training component implements different strategies. We compare: (1) the baseline learning strategy implemented in FairFace\cite{Karkkainen2021FairFace}, which jointly learns the feature extractor $\mathcal{F}$ and classifier $\mathcal{C}$, (2) the learning strategy implemented in DeepFace~\cite{serengil2024benchmark}, which fine-tunes the entire pretrained FR model, and (3) training only the classifier $\mathcal{C}$ on top of the frozen $\mathcal{F}^p$. Finally, we follow common practice to evaluate the accuracy and fairness of the obtained models, and add a robustness evaluation employing a newly introduced metric.}
    \label{fig_pipeline}
\end{figure*}

We first motivate the need for reliable DAI pipelines by theoretically analyzing the effect of accuracy in downstream face analysis tasks whose fairness needs evaluation.
This analysis is motivated by the observation that, for a given model, most existing fairness metrics do not correlate with DAI accuracy. 
For instance, if applied to a sufficiently large and balanced test dataset, a random classifier obtains a nearly perfect degree of bias scores since the accuracy of all demographic segments is almost identical. 
However, we demonstrate that the DAI accuracy has significant implications for estimating the fairness model in addressing a downstream task. 
Then, we present the main components of the DAI pipeline, summarized in \autoref{fig_pipeline}, namely the training datasets, DAI model training approaches, and evaluation dimensions. 

\subsection{Theoretical Motivation}\label{sec:theory}
We consider a face analysis system that performs face recognition (FR), face verification (FV), or other similar tasks. 
It is trained on an image dataset annotated with task-dependent labels, e.g., a class number associated with each image for recognition or a binary value for FV image pairs. 
These annotations are used to estimate the performance of the face analysis system~\cite{raji2020about_face}, with metrics such as micro-average accuracy, True Match Rate (TMR), or False Match Rate (FMR) (see~\autoref{sec:supp_metrics} in the appendix for details).

Importantly, each image also has sensitive attribute annotations, such as the age, gender, or ethnicity of the person represented. 
These attributes are required to estimate the fairness with metrics such as the {Degree of Bias (DoB)} ~\cite{gong2020joint_debiaising}, the Demographic Parity Difference (DPD) and {Demographic Parity Ratio (DPR)} \cite{equlizeodddemographocparity, demographicparity}, or the Equalized Odds Difference (EOD) and {Equalized Odds Ratio (EOR)}~\cite{equlizeodddemographocparity,equlizedodd}. 
Interestingly, most, if not all, of these fairness metrics can be expressed from a binary indicator $Y_i$, allowing us to derive a theoretical framework which is valid regardless of the metric further used (see details in~\autoref{sec:link_Yi_metrics} in the appendix).

In large-scale face datasets, annotating such attributes is tedious; thus, one often relies on automatic attribute classifiers to estimate them. These annotations are therefore prone to errors, which may further influence the quality of the fairness evaluation. 
To our knowledge, existing works do not address this point. 
In the following, we establish a plug-in estimator of the performance indicator conditioned on the \textit{estimated} sensitive attribute. We establish an upper bound on its bias relative to the true sensitive attribute as well as the expression of the covariance matrix of the unbiased estimator. Globally, this clarifies how a more accurate estimation of sensitive attributes enhances the fairness estimation of the face analysis system.

Let $G\in\{1,\dots,K\}$ be the true latent group of an identity, $\widehat G\in\{1,\dots,K\}$ the estimated label of the corresponding sensitive attribute, and $Y\in\{0,1\}$ a binary performance indicator associated to an identity for a given automatic FAS. For instance, $G$ could be the group \texttt{old$ \times$ caucasian $\times$ male} or \texttt{young $\times$ indian $\times$ female}. 
$Y$ is a generic FAS performance indicator that can be specialized to reflect various metrics such as the False Match Rate (see appendix~\autoref{sec:link_Yi_metrics}).

We define the prior of the sensitive attribute $a$ as  $\pi_a := \Pr(G=a)$ and the probability of success of the FAS given the \textit{true} said sensitive attribute as $p_a := \Pr(Y=1\mid G=a)$. This last is actually estimated automatically with some attribute classifiers that have a confusion matrix $C=(c_{ab})_{a,b=1}^K$ with $c_{ab} := \Pr(\widehat G=b \mid G=a)$ s.t each row of $C$ sums to 1. We then define $\tau_b := \Pr(\widehat G=b) = \sum_{a=1}^K \pi_a c_{ab}$ and denote the vectors $p=(p_1,\dots,p_K)^\top$, $\pi=(\pi_1,\dots,\pi_K)^\top$, and $\tau=(\tau_1,\dots,\tau_K)^\top=C^\top\pi$. We assume that, conditional on the true group $G$, the estimated label $\widehat G$ gives no further information about $Y$, that is:
\begin{equation}\label{eq:assume_conditinal_inped}
    (Y \perp\!\!\!\perp \widehat G)\mid G
\end{equation}
This denotes the conditional independence of $\widehat G$ and $Y$ given $G$. 
Said another way, we assume that the performance of the FAS depends on the identity and attribute group $G$ but not on the internal randomness of the attribute classifier that predicts $\widehat G$ instead of $G$. We also assume $C^\top$ is invertible (so $C$ is invertible). We consider an i.i.d. sample of $I$ identities; for identity $i$ we observe $(\widehat G_i,Y_i)$ but not $G_i$. For each observed label $g$ we define the empirical plug-in estimator
\begin{equation}    
\widehat m_g = \frac{\frac{1}{I}\sum_{i=1}^I \mathbf{1}\{\widehat G_i=g\} Y_i}{\frac{1}{I}\sum_{i=1}^I \mathbf{1}\{\widehat G_i=g\}}
\end{equation}

We note  $\widehat m=(\widehat m_1,\dots,\widehat m_K)^\top$ assuming  the denominator is nonzero for the $K$ groups considered, and $\widehat \tau$ the empirical estimator of $\tau$.  Hence:

\begin{restatable}{prop}{bias}
\label{theorem:bias}
(informal) Under the assumptions above, the empirical plug-in estimator $\widehat m_g$ converges in probability toward $m_g := \Pr(Y=1\mid \widehat G = g)$ as $I\to\infty$ and $m_g=\frac{1}{\tau_g}\sum_{a=1}^K\pi_ac_{ag}p_a$. Moreover, the bias of the plug-in limit relative to the true group rate $p_g$ satisfies

\begin{equation}
\mathrm{Bias}_g := m_g - p_g
= \frac{\sum_{a\ne g} \pi_a c_{a g} (p_a-p_g)}{\sum_{a=1}^K \pi_a c_{a g}},
\end{equation}

and therefore if $c_{ag}\to 0$ for every $a\ne g$ and $c_{g g}\to 1$ then $\mathrm{Bias}_g\to 0$. Moreover, for $\Delta_{\max,g} := \max_{a\ne g} |p_a-p_g|$,

\begin{equation}
    \big|\mathrm{Bias}_g\big|
\le \Delta_{\max,g}\cdot \frac{\sum_{a\ne g}\pi_a c_{a g}}{\pi_g c_{g g}}.
\end{equation}
\end{restatable}
Hence, if the confusion matrix tends towards identity -- that is, the estimation of the attributes is better -- the bias tends towards zero. 
Regarding the variance, we have: 

\begin{restatable}{prop}{covariance}
\label{theorem:covariance}
(informal) Assuming C is invertible, the estimator
\begin{equation}
\widehat p^{\mathrm{(corr)}} \;=\; \frac{(C^\top)^{-1} (\widehat\tau\odot\widehat m)}{\pi}
\end{equation}
is asymptotically unbiased for $p$. Moreover the multiplicative increase in variance is controled by the factor $\|(C^\top)^{-1}\|_{\mathrm{op}}^2$, where $\|\cdot\|_{\mathrm{op}}$ is the operator norm, such that when $C\to I$ the factor tends to $1$, while ill-conditioned $C$ (small singular values) amplify the variance.
\end{restatable}
The formal propositions and their proofs are in the ~\autoref{sec:sup_theory_proof}. The combination of both propositions shows that a better attribute inference leads to a less biased estimation of the metrics used to estimate the fairness of the FAS, with a smaller variance.

\subsection{Training datasets}

\label{subsec_dataset}
\autoref{fig_pipeline} provides a generic view of the DAI pipeline.
The curated dataset $\mathcal{D}$ yields a set $S = {s_1, s_2, ..., s_n}$ of demographic segments for each attribute.
For instance, ethnicity may be defined using predefined categories such as \textit{Asian, Black, Indian, White}~\cite{Karkkainen2021FairFace,wang2021meta}, country of origin~\cite{li2025instance,popescu2022face}, or skin appearance~\cite{merler2019diversity,Thong2023BeyondSkinTone}.
Any segmentation scheme introduces ethical and technical challenges~\cite{gwilliam2021rethinking,khan2021one,popescu2022face}, as it reduces the diversity of human faces into a finite set of $n$ classes.
In practice, we argue that segmentation should be selected to optimize for classification accuracy, fairness, and robustness jointly.
Beyond the segmentation choice itself, the dataset should contain diverse and sufficiently large samples for each segment, with reliable and high-quality labels~\cite{wang2021meta}.

\subsection{Model training}
\label{subsec_model}
Surprisingly, many widely used DAI models ($\mathcal{M}$), comprising a feature extractor $\mathcal{F}$ and a classifier $\mathcal{C}$, still rely on training from scratch~\cite{Karkkainen2021FairFace} or full fine-tuning~\cite{serengil2024benchmark} of pretrained models.
This persists despite the proven benefits of transfer learning in related face analysis tasks~\cite{narayan2025facexbench,sun2024face,amato2019face,fournier2025fairer}.
Motivated by these findings, we explore the use of pretrained models $\mathcal{F}^p$ as an alternative to full end-to-end training.
In the probing setting (\autoref{fig_pipeline}), we train the classifier $\mathcal{C}$ directly on top of a frozen backbone $\mathcal{F}^p$.
Prior DAI works~\cite{hunanyan2020race,kalkatawi2024ethnicity} that employed pretraining typically started from ImageNet-pretrained models.
However, due to the domain gap between general image classification and facial representations, this approach is suboptimal.
Instead, we use $\mathcal{F}^p$ pretrained for face recognition to leverage domain-relevant representations.
A common assumption~\cite{Karkkainen2021FairFace,sarridis2023towards} is that the demographic segments $S$ are linearly separable in the embedding space, motivating the use of a linear classifier $\mathcal{C}$.
We challenge this assumption and instead adopt non-linear classification heads, which we show to improve performance.

\subsection{Evaluation dimensions}
\label{subsec_metrics}
DAI evaluation typically involves accuracy metrics, often complemented by fairness indicators~\cite{sarridis2023towards}.
These allow per-segment behavior analysis and deeper performance diagnostics~\cite{fournier2025fairer}.
A key challenge is to balance accuracy and fairness, as improvements in one may degrade the other~\cite{zarei2025privacy}.
We include both in our evaluation and introduce robustness as a complementary dimension.

We posit that a reliable attribute classifier should produce consistent predictions for different images of the same identity.
In practice, this consistency is not guaranteed due to variations in illumination, pose, resolution, and other capture conditions~\cite{merler2019diversity,Thong2023BeyondSkinTone}, as illustrated in \autoref{fig_teaser}.

We define a robustness metric below, which is \textbf{taxonomy agnostic}, since it leverages not Ethnicity but Identity label.
Let us consider an image dataset $\mathcal{I}=\{I^{id}_j; id\in [\![1,D]\!, j\in [\![1,N_{id}]\!] \}$, with $D$ identities, an identitity $id$ being represented by $N_{id}$ images. 
We consider a labeling function $f_\mathcal{A}^\mathcal{C}$ that attributes a label to each image, each label being an element of a finite set $\mathcal{C}$, and depends on a dataset $\mathcal{A}$ ($f_\mathcal{A}^\mathcal{C}$ is defined below). 

For each identity $id$, we compute the relative frequency of each possible label $k\in\mathcal{C}$ noted as $p_{id}^k = \frac{1}{N_{id}} \sum_{j=1}^{N^{id}} \delta_{f_\mathcal{A}^\mathcal{C}(I^{id}_j)=f_\mathcal{A}^\mathcal{C}(I^{id}_k)}$. We then compute the \textbf{homogeneity entropy} as the average entropy of $p$ over all identities:
\begin{equation}\label{eq:def_HomE}
HomE = \frac{1}{D} \sum_{id}\frac{1}{\log(C)}\sum_k - p_{id}^k \log(p_{id}^k)   
\end{equation}

This entropy-based measure captures intra-identity consistency.
Lower values of \texttt{HomE} indicate higher robustness and are thus desirable for trustworthy DAI pipelines.

\section{Experiments}
\label{sec:expe}

We first present the experimental setup used to audit demographic attribute inference.
Next, we report and analyze the main results obtained with both the baseline and the proposed DAI pipelines.
Finally, we present ablation experiments to study the contribution of the main components of our approach.

\subsection{Experimental setup}

\paragraph{DAI pipelines.} 
We compare our probing-based demographic attribute inference approach against two popular baselines: FairFace~\cite{Karkkainen2021FairFace} and DeepFace~\cite{serengil2024benchmark}.
FairFace uses a ResNet34 (RN34)~\cite{he2016deep} architecture and combines a publicly released subset with proprietary data that is not available~\cite{Karkkainen2021FairFace}.
We also train a DAI pipeline with the FairFace public subset and the ResNet34 architecture in an attempt to reproduce the original results. 
The FairFace baselines implement the full training pipeline illustrated in \autoref{fig_pipeline}. 
DeepFace~\cite{serengil2024benchmark} performs full fine-tuning on top of a VGGFace model~\cite{parkhi2015deep} pretrained for face recognition. It combines training data from multiple sources and follows the fine-tuning procedure shown in \autoref{fig_pipeline}.

Our probing-based DAI pipelines use a variety of pretrained face recognition (FR) models combined with an RBF classifier.
We evaluate both CNN and transformer architectures with a comparable parametric footprint to the baselines.
Specifically, we test ResNet34, ResNet50, iResNet18 (IR18)~\cite{deng2019arcface}, and ViT-T~\cite{dosovitskiy2021image}, whose parameter counts range from 19.1M to 24.6M.
To explore the effect of increasing model size, we additionally use ResNet101 (RN101), iResNet50 (IR50), and ViT-S~\cite{dosovitskiy2021image}.
We use the InsightFace implementations for transformer-based models\footnote{\url{https://github.com/deepinsight/insightface}}.
The pretrained models are trained on a subset of the WebFace260M dataset~\cite{zhu2023webface260m}, comprising approximately 28k identities and 1 million images.
We selected this subset size based on GPU constraints.
To assess the impact of pretraining data size, we also experiment with existing AdaFace~\cite{kim2022adaface} models trained on the full WebFace dataset. 

We implement two types of classification heads: (1) an RBF-kernel SVM (denoted RBF in \autoref{tab_results}) for non-linear probing, and (2) a linear SVM (denoted LIN in \autoref{tab_results}) for linear probing.
Both classifiers are trained on a balanced subset of 100k images from BUPT-Balanced, covering 25.3k identities and balanced across gender and ethnicity.
This image count matches that of the FairFace public subset~\cite{Karkkainen2021FairFace} for fair comparison.
Ethnicity labels are directly available in BUPT.
For gender, we leverage the presence of multiple samples per identity and use majority voting over predictions made by FairFace.
This strategy, evaluated on a 500-identity subset, yields gender label accuracy exceeding 99.5\%, making it reliable for training and evaluation.
We reserve a separate BUPT validation subset containing 60k images and 1,330 identities.
All reported results use the DAI pipeline versions that maximize accuracy on this validation set.

\begin{table*}[ht]
\centering\centering
\resizebox{\linewidth}{!}{
\begin{tabular}{l|c|c|ccc|ccc|cc|cc|cc|cc}
\multirow{3}{*}{\large DAI Pipeline} & \multirow{3}{*}{\rotatebox{60}{\large \#param (M)}} & \multirow{3}{*}{\rotatebox{60}{\large \#imgs (M)}} 
& \multicolumn{8}{c|}{\large Ethnicity} 
& \multicolumn{6}{c}{\large Gender} \\
\cline{4-17}
& & & \multicolumn{3}{c|}{\large Accuracy $\uparrow$} & \multicolumn{3}{c|}{\large Fairness $\downarrow$} & \multicolumn{2}{c|}{\large  Robustness $\downarrow$} 
& \multicolumn{2}{c|}{\large Accuracy $\uparrow$} & \multicolumn{2}{c|}{\large Fairness $\downarrow$} & \multicolumn{2}{c}{\large Robustness $\downarrow$} \\
\cline{4-17}
& & & \rotatebox{60}{BFW} & \rotatebox{60}{RFW} & \rotatebox{60}{BUPT} & 
\rotatebox{60}{BFW} & \rotatebox{60}{RFW} & \rotatebox{60}{BUPT} & 
\rotatebox{60}{BFW} & \rotatebox{60}{BUPT} & 

\rotatebox{60}{BFW} &  \rotatebox{60}{BUPT} & 
\rotatebox{60}{BFW} &  \rotatebox{60}{BUPT} & 
\rotatebox{60}{BFW} & \rotatebox{60}{BUPT} \\
\hline
\rowcolor{red!15}
FairFace~\cite{Karkkainen2021FairFace}   & 21.3 & - & 87.87 & 88.08 & 82.27 & 3.50 & 4.10 & 4.58 & 0.19 & 0.22 
& 96.27  &  98.05 &  2.01 & \underline{\textit{\textbf{1.31}}} & 0.19 & 0.22 \\
\rowcolor{red!15} FairFace$_{R}$ & 21.3 & - & 64.04 & 67.53 & 59.55 & 9.83 & 9.76 & 10.62 & 0.38 & 0.41 & 92.27 & 95.21 & 2.29 & 4.17  &  0.25 & 0.19\\
\rowcolor{red!15} FairFace$_{B}$ & 21.3 & - & 78.22 & 77.83 & 89.41 & 4.95 & 6.89 & 2.28 & 0.35 & 0.18 & 93.80 & 96.69 & 2.43 & 4.20  &  0.19 & 0.12\\

\rowcolor{red!15} DeepFace\cite{serengil2024benchmark}& 134 & - & 82.29 & 82.31 & 74.37 & 1.72 &  3.15 & 3.03 & 0.29 & 0.36
& 85.43   & 91.22 & 26.88 &  25.55 & 0.31  & 0.16\\
\rowcolor{green!15} RN34+RBF     & 21.3 & 1 & 92.42 & 91.78 & 88.69 & 1.96 & 2.50 & 2.72 & 0.14 & 0.17  
& 95.48 & 97.23  & \underline{\textit{\textbf{0.04}}} & 1.93 & 0.16 & 0.11 \\
\rowcolor{green!15} IR18+RBF         & 24.0 & 1 & 93.39 & 92.94 & 90.51 & 1.65 & 1.82 & 2.38 & 0.13 & 0.14 
&  96.68  & 98.06 & 1.22  &  2.95  & 0.14 & \textit{0.08} \\
\rowcolor{green!15} RN50+RBF         & 24.6 & 1 & \textit{94.99}  & \textit{94.09} & 91.62 & 1.48 & 1.69 & 2.20 & \textit{0.09} & 0.11 
& \textit{97.51}   & 98.30  & 0.18 & 1.70 & \underline{\textit{\textbf{0.12}}} & \textit{0.08} \\
\rowcolor{green!15} ViT-T+RBF        & 19.1 & 1 & 94.72  & 94.01 & \textit{91.67} & \textit{1.33} & \textit{1.44} & \textit{2.14} & 0.10 & \textit{0.10} 
&  97.09 &  \textit{98.38} & 0.29  & 1.80  & 0.14 & \textit{0.08} \\
\rowcolor{blue!15} RN101+RBF         & 43.5 & 1 & 95.00  & 94.09 & 91.31 & 1.45 & 1.79 & 2.27 & 0.09 & 0.11 
&  97.61 &  98.35 & 0.55 & 1.57  & 0.13 & 0.08  \\
\rowcolor{blue!15}  IR50+LIN      & 43.5  & 1 & 90.61 & 90.65 & 87.95 & 2.14 & 2.58 & 2.67 & 0.19 & 0.20 & 94.96 & 96.85 & 1.55 & 1.32  & 0.19 & 0.13 \\
\rowcolor{blue!15} IR50+RBF         & 43.6 & 1 & 94.98 & \underline{\textbf{95.21}} & \underline{\textbf{92.5}} & \underline{\textbf{1.19}} & 1.28 & 2.16  & \underline{\textbf{0.08}} & \underline{\textbf{0.09}}
& \underline{97.74}  & \underline{98.59} & 0.40 & 1.91  & \underline{\textbf{0.12}} & \underline{\textbf{0.07}} \\
\rowcolor{blue!15} ViT-S+RBF        & 76.0 & 1 & \underline{95.48} & 94.77 & 92.06 & 1.21 & \underline{\textbf{1.27}} & \underline{2.05} & 0.11 & \underline{\textbf{0.09}} 
& 97.33 & 98.53 & 0.42 & 1.84  & 0.13 &  \underline{\textbf{0.07}} \\

\rowcolor{orange!15} IR18+RBF$_{pt}$       & 24.0 & 40 & 94.49 & 94.00 & 91.85 & 1.39 & 1.53 & 2.10 & 0.13 & 0.14 & 97.09 & 98.39 & 0.19 & 2.00  & 0.13 & 0.08 \\
\rowcolor{orange!15}    IR50+RBF$_{pt}$  & 43.6 & 40 & \textbf{95.55} & 95.01 & 92.36 & 1.26 & 1.42 & \textbf{2.02} & 0.18 & 0.16 & \textbf{97.83} & \textbf{98.72} & 0.6 & 1.65 & 0.13 &  0.08    \\

\end{tabular}
}

\caption{Comparison of demographic attribute inference pipelines based on accuracy, fairness, and robustness dimensions. Baselines are color-coded in red. Column 3 is the number of pretraining images (in millions). FairFace$_{R}$ and FairFace$_{B}$ indicate unsuccessful attempts to reproduce FairFace with their public dataset and the BUPT 100k training subset. DAI pipelines pretrained here with 1M WebFace images are color-coded in green when the parametric footprint is similar to that of FairFace and in blue for larger backbones. DAI pipelines using preexisting models pretrained with 40M WebFace images~\cite{zhu2023webface260m} ($_{pt}$ suffix) are color-coded in orange. RBF and LIN designate the non-linear and linear classification heads when using pretraining. The parameter counts (\#param) of the backbones are estimated with Torch Lightning for all pipelines to avoid any inconsistencies. The best results for compact models comparable with FairFace~\cite{Karkkainen2021FairFace} (red and green rows) are in \textit{italics}. The best results obtained when including larger pretrained architectures (red, green, and blue rows) are \underline{underlined}. The best global results are in \textbf{bold}. Higher is better for accuracy, while lower is better for fairness and robustness.}
\vspace{-4mm}
\label{tab_results}
\end{table*}

\paragraph{Test datasets.} 
We evaluate the reliability of ethnicity and gender inference using publicly available datasets selected based on the availability of demographic attributes and identity labels.
BFW~\cite{bias3}, RFW~\cite{Wang2019RFW}, and BUPT~\cite{wang2021meta} all employ a common four-class ethnicity segmentation and are used for accuracy and fairness evaluation.
We sample 63k images from BUPT (1,400 identities) balanced for ethnicity.
As BUPT is also used for DAI training and validation, we ensure that this test subset is completely disjoint from the training and validation sets (disjoint set of Identities).

BFW and BUPT include multiple samples per identity and are used for evaluating ethnicity robustness.
BFW includes gender annotations, enabling evaluation across all three dimensions.
BUPT gender labels are inferred using the same identity-based majority voting procedure applied during classifier training, ensuring consistency.

\paragraph{Metrics.}
We report micro-average for accuracy evaluation.
For fairness, we use the degree of bias (DoB), computed as the standard deviation of per-group accuracies (lower is better).
For robustness, we use the Homogeneity Entropy (HomE) metric defined in \autoref{eq:def_HomE}, along with additional robustness measures described in \autoref{sec:supp_metrics}.

\subsection{Main results}
\label{subsec_main_res}

\autoref{tab_results} presents the accuracy, fairness, and robustness scores for both ethnicity and gender inference pipelines.
Overall, the proposed approach achieves substantial improvements across all evaluation dimensions and datasets.

\paragraph{Ethnicity.} 
 The accuracy gains range from 7 to 10 percentage points, depending on the dataset.
Compared with FairFace~\cite{Karkkainen2021FairFace}, the degree of bias (DoB) is reduced by a factor of 2–3, and HomE is approximately halved.
The comparison with DeepFace~\cite{serengil2024benchmark} is even more favorable in terms of accuracy and robustness, although DeepFace exhibits better fairness scores than FairFace despite lower overall accuracy.
Interestingly, performance differences between compact and larger models remain modest when using a pretrained FR model with an RBF classifier—highlighting the robustness and efficiency of this simple yet effective DAI pipeline.
Among compact architectures, RN34 yields the weakest results, while RN50 and ViT-T perform comparably.
For larger backbones (highlighted in blue), IR50 and ViT-S achieve slightly better accuracy and show notable improvements in fairness and robustness.
The comparison between IR50+RBF and IR50+LIN underscores the importance of the classification head.
The performance gain with RBF suggests that the pretrained encoder does not linearly separate the four ethnicity classes—indicating that a non-linear classifier is essential for effective inference.
Pretraining with a larger dataset (orange cells) does not consistently improve accuracy or fairness and tends to reduce robustness.
The main difference lies in dataset structure: the 1M-image WebFace subset is balanced, whereas the 40M-image version is highly imbalanced.
These results indicate a trade-off between dataset size and class balance, which we further investigate in \autoref{subsec_ablation}.
In practice, we recommend RN50+RBF or ViT-T+RBF for computationally constrained environments and
IR50+RBF for high-performance settings.

\paragraph{Gender.} 
This inference is simpler than ethnicity inference, as previously noted~\cite{Karkkainen2021FairFace} and confirmed by results in \autoref{tab_results}.
The proposed pipelines show consistently strong performance, confirming the stability of the transfer-learning-based approach.
Although performance differences are smaller due to task saturation, IR50+RBF—our best-performing configuration—still improves accuracy by 1.5 points on BFW and 0.5 points on BUPT compared to FairFace.
In terms of fairness, our method improves performance on BFW, but FairFace remains fairer on BUPT.
Robustness improves consistently across both datasets relative to FairFace.
Among compact models, RN50+RBF performs best, with ViT-T+RBF as a close second.
For larger backbones, IR50+RBF slightly outperforms ViT-S+RBF, although the differences are minor.
As with ethnicity, larger models provide modest gains over compact ones.
Pretraining with the larger WebFace dataset slightly improves accuracy but produces mixed effects on fairness and robustness.

Considering both ethnicity and gender performance, RN50+RBF and IR50+RBF emerge as the most reliable DAI pipelines among compact and large architectures, respectively.
In practice, the choice between them depends on computational budget and deployment constraints. 

\subsection{Ablation and analysis}
\label{subsec_ablation}
We perform two ablations for ethnicity inference to highlight the impact of imbalance in the pretraining dataset and that of size and structure in the BUPT subset used to train the ethnicity classifier.
We then present the detailed accuracy and robustness results for ethnicity in RFW.

\paragraph{Pretraining dataset balancing.}
WebFace over-represents White individuals, while the pretraining subset used in~\autoref{subsec_main_res} is balanced.
We sample 1M WebFace images randomly to assess the impact of imbalance during pretraining.
\autoref{tab:best_accuracy_unbalanced_balanced} presents the accuracy obtained with the IR50+RBF model.
The results indicate that balancing during pretraining generally provides a slight advantage when inferring ethnicity. 

\begin{table}[ht]
\centering

\resizebox{0.8\linewidth}{!}{
\begin{tabular}{lcccccc
}
\toprule
\textbf{Arch.} & \multicolumn{2}{c}{\textbf{BFW}} & \multicolumn{2}{c}{\textbf{RFW}} & \multicolumn{2}{c}{\textbf{BUPT}} \\ 
\cmidrule(lr){2-3} \cmidrule(lr){4-5} \cmidrule(lr){6-7}
& {Imb} & {Bal} & {Imb} & {Bal} & {Imbb} & {Bal} \\
\midrule
IR-18    & 92.7 & \bfseries 93.2 & 92.6 & \bfseries 93.0 & \bfseries 90.5 & 90.4 \\
IR-50    & 95.3 & \bfseries 95.6 & 94.8 & \bfseries 95.2 & 91.8 & \bfseries 92.5 \\
RN101    & \bfseries 94.0 & 93.8 & \bfseries 93.4 & 93.4 & 90.8 & \bfseries 91.1 \\
RN34     & 89.8 & \bfseries 91.4 & 88.4 & \bfseries 90.5 & 86.9 & \bfseries 87.9 \\
RN50     & 94.1 & \bfseries 95.0 & 93.5 & \bfseries 94.7 & 90.8 & \bfseries 91.6 \\
ViT-s    & 94.5 & \bfseries 95.5 & 94.0 & \bfseries 94.8 & 91.4 & \bfseries 92.1 \\
ViT-t   & 93.7 & \bfseries 94.7 & 93.4 & \bfseries 94.0 & 91.3 & \bfseries 91.7 \\
\bottomrule
\end{tabular}
}
\caption{BFW, RFW, and BUPT accuracy when pretraining the IR50+RBF encoder with Imbalanced (Imb) or balanced (Bal) WebFace subset comprising 1M images.}
\vspace{-3mm}
\label{tab:best_accuracy_unbalanced_balanced}
\end{table}

\paragraph{Ethnicity classification dataset size and structure.}
\autoref{tab_IR_50_cross_table} details the results obtained with different sizes and structures of the BUPT subset used to train the ethnicity classification head. 
The results on the individual rows and columns indicate that a larger training dataset usually improves accuracy. 
Going beyond 100,000 training samples would probably bring further slight improvement, but such an experiment is outside the immediate scope of our study. 
The comparison of identity-sample combinations totaling the same number of images indicates a systematic advantage for configurations including more identities and a lower sample count per identity. 
This finding hints at the usefulness of maximizing diversity in DAI training datasets, to limit the number of sensitive attributes they contain.

\begin{table}[ht]
\centering
\small 

\begin{tabular}[bt]{lccccccc
}
\toprule
 &
{\textbf{1}} & {\textbf{2}} & {\textbf{5}} & {\textbf{10}} & {\textbf{15}} & {\textbf{20}} & {\textbf{40}} \\
\midrule
500 & 85.2 & 87.3 & 88.4 & 89.4 & 89.2 & 89.5 & 89.6 \\
1000 & 87.8 & 89.8 & 90.3 & 90.6 & 90.8 & 90.8 & 91.0 \\
2000 & 89.6 & 90.2 & 90.8 & 91.3 & 91.6 & 91.4 & 92.1 \\
5000 & 90.9 & 91.6 & 91.8 & 93.7 & 94.2 & 94.3 & {---} \\
10000 & 91.7 & 91.5 & 94.3 & 94.8 & {---} & {---} & {---} \\
20000 & 92.0 & 94.3 & 95.2 & {---} & {---} & {---} & {---} \\
\bottomrule
\end{tabular}
\caption{Impact of the structure of the ethnicity classification dataset. We vary the number of identities (rows) and the number of samples per identity (columns). 40 is the maximum number of samples per identity. We limit the total dataset size to 100k. We report the RFW accuracy obtained with the IR50+RBF backbone pretrained on 1M WebFace images. }
\vspace{-3mm}
\label{tab_IR_50_cross_table} 
\end{table}

\begin{figure}[ht]
    \centering
    \includegraphics[width=\linewidth]{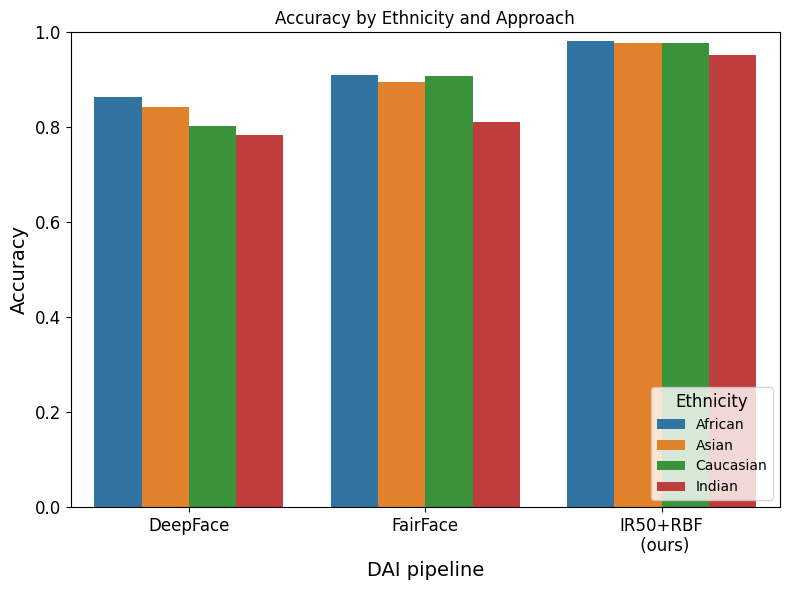}
    \caption{Ethnicity segments accuracy in RFW for the two main baselines and IR50+RBF pretrained with 1M WebFace images.}
    \label{fig_detail_acc}
\end{figure}

\begin{figure}[ht]
    \centering
    \includegraphics[width=\linewidth]{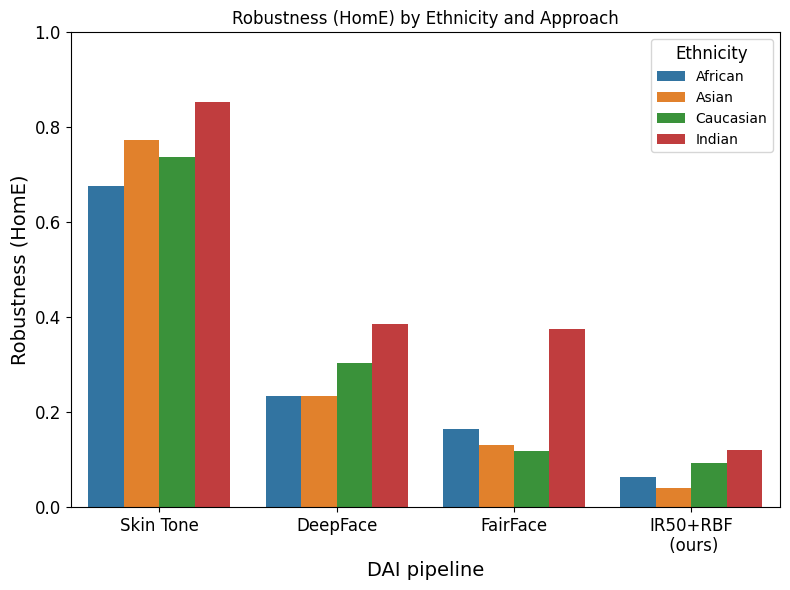}
    \caption{Ethnicity segments robustness (lower is better) in RFW for the two main baselines, IR50+RBF pretrained with 1M WebFace images, and with four skin-color-based segments from~\cite{Thong2023BeyondSkinTone}. }
    \vspace{-3mm}
    \label{fig_detail_rob}
\end{figure}

\paragraph{RFW ethnicity inference detail.}
\autoref{fig_detail_acc} and~\autoref{fig_detail_rob} detail the accuracy and robustness distributions per ethnicity. 
They underline the improved accuracy of the proposed IR50+RBF model and the lower performance variability across segments compared with the baselines.

The robustness per ethnicity (\autoref{fig_detail_rob}) is lower and more stable for IR50+RBF compared with FairFace and DeepFace, the two ethnicity-based baselines.
The robustness gain is even bigger compared with~\cite{Thong2023BeyondSkinTone}. 
This skin-based appearance segmentation exhibits poor robustness globally, and scores vary significantly across segments. 
This last comparison exemplifies the applicability of the robustness metric introduced in~\autoref{eq:def_HomE} to different segmentation schemes. 
Our finding questions the practical utility of the metric introduced in~\cite{Thong2023BeyondSkinTone} since it is unable to predict the same segment for different photos of the same identity. 

\section{Limitations and Ethics}
\textbf{Ethnicity segmentation is coarse and limited.}
The use of predefined ethnicity categories simplifies the inference but fails to capture the complex, continuous, and intersectional nature of human phenotypic diversity.
In this work, we adopt a four-class scheme (Asian, Black, Indian, White).
While commonly used and empirically robust, it reduces rich demographic variation into coarse groupings.
This simplification may obscure subtle patterns of bias or robustness within subgroups and may limit generalizability across cultural or geographic contexts.
Designing segmentation schemes that are both ethically meaningful and statistically reliable remains an open challenge, requiring careful trade-offs between interpretability, granularity, and practicality.

\textbf{Model size and pretraining data: larger may still be better.}
Our results show that transfer learning from compact face recognition models yields strong performance.
However, we also observe modest but consistent gains when using larger architectures and more extensive pretraining datasets.
This suggests that current performance may be suboptimal and that further scaling could improve outcomes.
Nonetheless, training or fine-tuning large models introduces computational and environmental costs, and large-scale datasets often lack transparency regarding their demographic composition.
Balancing scalability, efficiency, and dataset accountability is increasingly important in the era of foundation models.

\textbf{Limited interpretability of learned demographic features.}
While our approach improves performance across accuracy, fairness, and robustness, the internal representations learned by deep models remain largely opaque.
This lack of interpretability hampers transparency in high-stakes applications, making it difficult to audit which features influence predictions.
Improving interpretability without sacrificing accuracy remains an important research direction, particularly as current architectures prioritize representational abstraction over explainability.

\textbf{Ethical risks and potential misuse.}
Although our models are designed for fairness auditing and scientific reproducibility, they could be repurposed for unethical uses such as surveillance, profiling, or discrimination.
The ability to infer sensitive demographic attributes from facial images at scale raises significant concerns around consent, privacy, and bias amplification.
While releasing models and code promotes transparency and peer scrutiny, technical safeguards alone are insufficient.
A broader dialogue is needed within the research, ethics, and policy communities to define acceptable use cases, auditing frameworks, and accountability mechanisms.

\section{Conclusion}
We revisited the DAI process and proposed a transfer learning–based formulation, supported by a theoretical motivation for improving the reliability of DAI pipelines.
We extended the evaluation framework beyond accuracy and fairness by introducing robustness as a third critical dimension, and proposed HomE, an entropy-based robustness metric grounded in intra-identity consistency.
We implemented and evaluated the proposed approach for both ethnicity and gender inference, demonstrating consistent performance gains—particularly for ethnicity, which remains the more challenging attribute.
These results were obtained with a simple yet effective probing-based design, showing that a modular and interpretable architecture can match or surpass existing end-to-end systems.

Looking forward, we will address the limitations identified above.
First, refining ethnicity categorization remains an open challenge, as shown by the limited robustness of recent schemes such as~\cite{Thong2023BeyondSkinTone}.
We will explore the construction of datasets organized around finer-grained ethnic categories with coherent phenotypic traits, as well as unsupervised, data-driven segmentations based on clustering embeddings from large pretrained models.
Second, we plan to incorporate new foundation models as they emerge, evaluating their impact on DAI performance and robustness.
Third, we will investigate techniques for interpreting demographic embeddings, such as concept activation vectors and disentanglement-based methods.
We also aim to explore whether integrating explainability constraints during training can improve model transparency without compromising predictive performance.

\paragraph*{Acknowledgement} This work was partly supported by the SHARP ANR project ANR-23-PEIA-0008 in the context of the France 2030 program. It was provided with computer and storage resources by GENCI at IDRIS thanks to the grant 2024-AD011014868R1 on the supercomputer Jean Zay’s the CSL, H100 and A100 partition.

{
    \small
    \bibliographystyle{ieeenat_fullname}
    \bibliography{main}
}

\clearpage
\appendix

\section{Implementation detailled description}\label{sec:supp_datasets}

\subsection{Datasets}
FairFace~\cite{Karkkainen2021FairFace} is described in detail in~\autoref{sec:related}.

BUPT~\cite{wang2021meta} is a dataset of real images that can be found \href{http://www.whdeng.cn/RFW/index.html}{at this URL}. We actually consider \textit{BUPT-BalanceFace} that contains 1.3M images from 28k celebrities. 
Four labels are considered, namely \textit{caucasian}, \textit{indian}, \textit{asian} and \textit{african}. The dataset is balanced w.r.t the number of identity per class (7k per label) and approximatelly balanced in terms of number of images: 324k for \textit{african}, 325k for \textit{asian}, 326k for \textit{caucasian} and 375k for \textit{indian}.

WebFace~\cite{zhu2023webface260m} is a dataset of real images, containing 4M identities represented by 260M face images, which a part (4M identities and 42M face images) has been curated to result into Webface42M.

\subsection{Training details}

\textbf{For training the end-to-end classifier} (i.e., FairFace$_{R}$ and FairFace$_{B}$), we set the training duration to 30 epochs with a batch size of 64. We used the SGD optimizer with momentum coupled with a OneCycle LR scheduler strategy, starting with a learning rate of 0.001 that increases linearly during the first third of the total epochs to reach 0.1. The learning rate then decreases to 0 using cosine annealing. This approach outperforms the original FairFace authors' use of Adam with a constant learning rate. For data augmentation during training, we applied random horizontal flipping, low-resolution augmentation, and random centered cropping with the same parameters as those in the AdaFace~\cite{kim2022adaface} repository. The CrossEntropy loss converges to 0.03 at the end of training, demonstrating successful convergence.

\noindent For training the \textbf{face recognition classifiers}, we employed the standard setup from AdaFace. We used the AdaFace loss with parameter h set to 0.333 and m set to 0.4. We applied the standard framework augmentations for face recognition training: random flipping, crop augmentation, low-resolution augmentation, and photometric augmentation. We trained the model for 30 epochs using the SGD optimizer with momentum and weight decay fixed at 5e-4. The scheduler was a step LR scheduler that divides the learning rate by 10 at epochs 12, 20, and 24. To ensure stability of the AdaFace loss when training the ResNets (torchvision implementation), we add a fully connected layer to the models to project the original 2024 latent space to 512.

\noindent \textbf{The SVM implementation} is from scikit-learn ~\cite{scikit-learn}. To accelerate the computation of cross distances, we implemented a wrapper that performs this phase on GPU. This allows us to drastically reduce both training and inference time (training time drops from 40 minutes to 2 minutes on a training set of 100k samples). For the RBF kernel, we use the scaling gamma defined as \(\frac{1}{D \cdot \text{Var}(X)}\), where \(D\) is the dimension of the feature vectors (512) and \(X\) are the input vectors. Since BUPT is balanced with regard to ethnicity but not gender, we use balanced weighting for gender defined as: \(\text{Weight}(\text{Class}_i) = \frac{0.5}{\text{Frequency}(\text{Class}_i)}\). We use the squared hinge loss with L2 penalization and 1.0 as the regularization parameter.

All trainings are performed on a compute device having 250 GB of RAM, 96 CPU cores and an NVIDIA H100 GPU.

\textbf{}

\section{Details on metrics}\label{sec:supp_metrics}

\subsection{Performance Metrics}
The face recognition performance is estimated with \textit{Micro-Average Accuracy} which is reelvant when dealing with unbalanced data, as it gives equal importance to each dataset segment, regardless its size. Therefore, the metrics is not biased toward the majority group. Other possible performance metrics include:
\begin{itemize}
    \item \textit{True Match Rate (TMR)} or \textit{True positive Rate (TPR)} or \textit{Sensitivity} is the number of actual positive cases correctly identified over the actual number of positive samples in the dataset 
    \item \textit{False Match Rate (FMR)}, or \textit{False Positive Rate (FPR)}, is the number of negative cases incorrectly identified as positive by the FAS over the actual number of negative samples.
\end{itemize}

\subsection{Fairness Metrics}\label{sec:supp_metrics_fairness}
To estiamte the fairness in the main paper, we considered the Degree of Bias (DoB) \cite{gong2020joint_debiaising} which is the standard deviation of the per-group accuracies. A lower standard deviation means that the  performance varies less from one group to another, thus that the FAS is fairer.

\textit{Demographic Parity Difference (DPD)} and \textit{Demographic Parity Ratio (DPR)} \cite{equlizeodddemographocparity, demographicparity} are derived from the same principle, aiming at achieving equal representation of the demographic groups. To compute them, one consider the lowest $p_{min}$ and highest $p_{max}$ probability across all subgroups then $DPD=|p_{max}-p_{min}|$ and $DPR=\frac{p_{min}}{p_{max}}$. Hence both are scores in $[0,1]$ but the lower DPD, the faire the FAS while it is fairer when DPR is closer to one. 

\textit{Equalized Odds Difference (EOD)} and \textit{Equalized Odds Ratio (EOR)} \cite{equlizeodddemographocparity,equlizedodd}  require that the FAS's performances  are independent of the demographic group, that is the TPR or FPR are the same across all groups. 
The EOD is defined by the absolute difference between smallest and largest values for the considered performance metric (TPR or FPR) across groups, while EOR is defined according to their ratio. Similarly to demographic parity, these matrics are in $[0, 1]$ a FAS is perfectly fair when its EOD is zero or its EOR is one.

We present a formal presentation of these metrics and their link to the binary indicator of performance ($Y_i$ in the main article) in \autoref{sec:link_Yi_metrics}.

\subsection{Label Robustness Metric}\label{sec:supp_metrics_robustness}
Let us consider an image dataset $\mathcal{I}=\{I^{id}_j; id\in [\![1,D]\!, j\in [\![1,N_{id}]\!] \}$, with $D$ identities, an identitity $id$ being represented by $N_{id}$ images. 
Such dataset are typically CASIA or Wikidiv. We consider a labeling function $f_\mathcal{A}^\mathcal{C}$ that attributes a label to each image, each label being an element of a finite set $\mathcal{C}$, and depends of a dataset $\mathcal{A}$ ($f_\mathcal{A}^\mathcal{C}$ is defined below). 
We propose to consider three metrics to reflect the robustness of the labeling system $(\mathcal{A},\mathcal{C})$. The general idea is that a good labeling system should (at least) attribute as much as possible images of a given identity to the same label.

\textbf{Macro majority accuracy}. For each identity $id$, we consider the  frequency of the majority label and compute the average over all labels of the dataset 
\begin{equation}
    MaMA = \frac{\displaystyle\sum_{id=1}^D  \displaystyle\max_{k=1...N_{id}} \left( \sum_{j=1}^{N^{id}} \delta_{f_\mathcal{A}^\mathcal{C}(I^{id}_j)=f_\mathcal{A}^\mathcal{C}(I^{id}_k)} \right)}{\sum_{id} N_{id}}
\end{equation}
    
\textbf{Micro majority accuracy} For each identity, we consider the proportion of images labeled with the majority label and compute its average over all identities: 
\begin{equation}
    MiMA = \frac{1}{D}\sum_{id=1}^D \frac{1}{N_{id}} \displaystyle\max_{k=1...N_{id}} \left( \sum_{j=1}^{N^{id}} \delta_{f_\mathcal{A}^\mathcal{C}(I^{id}_j)=f_\mathcal{A}^\mathcal{C}(I^{id}_k)}\right)
\end{equation}

Last, let remind that in the main paper we consider, for each identity $id$, the relative frequency of each possible label $k\in\mathcal{C}$ noted as $p_{id}^k = \frac{1}{N_{id}} \sum_{j=1}^{N^{id}} \delta_{f_\mathcal{A}^\mathcal{C}(I^{id}_j)=f_\mathcal{A}^\mathcal{C}(I^{id}_k)}$. We then compute the \textbf{homogeneity entropy} as the average entropy of $p$ over all identities:
\begin{equation*}
HomE = \frac{1}{D} \sum_{id}\frac{1}{\log(C)}\sum_k - p_{id}^k \log(p_{id}^k)    \tag{\ref{eq:def_HomE} revisited} 
\end{equation*}

These metrics reflect the average variation of the attributions for each entity. Ideally, all images of a given entity are attributed to the same label. In this case, the \textit{homogeneity entropy} \textit{macro majority accuracy} and \textit{micro majority accuracy}  are respectively 0,1 and 1. In the worst case, the images of a given entity are uniformly distributed over all $C$ labels, leading to the three metrics above being respectively 1, $\frac{1}{C}$ and $ \frac{1}{C}$. 
We thus normalize MiMA and MaMA by multiplying by $\frac{C}{C-1}$ and removing $\frac{1}{C-1}$ such that the worst value is 0 and the best value is 1

\textbf{Labelling function} the function $f_\mathcal{A}^\mathcal{C}$ project an image onto a unique label in $\mathcal{C}$. It is defined according an image dataset $\mathcal{A}$. We consider two versions of $f_\mathcal{A}^\mathcal{C}$:
\begin{itemize}
    \item a non-parametric function: each image of $\mathcal{A}$ is described with an image encoder that project it into an euclidean space. It then attributes a label to an image according to its nearest neighbors (1-NN or k-NN)
    \item a parametric function, such as a neural network, that is learnt on $\mathcal{A}$, with labels in $\mathcal{C}$, and a cross-entropy loss.
\end{itemize}

\subsection{Robustness Results}\label{sec:supp_results_robustness}
Results of robutness for all the methods are reported in \autoref{tab:supp_all_robustness}, in term of all three metrics.

\begin{table*}[ht]
\centering\centering
\resizebox{\linewidth}{!}{
\begin{tabular}{l|c|c|cc|cc|cc|cc|cc|cc}
\multirow{3}{*}{\large DAI System} & \multirow{3}{*}{\rotatebox{60}{\large \#param (M)}} & \multirow{3}{*}{\rotatebox{60}{\large \#img (M)}} 
& \multicolumn{6}{c|}{\large Ethnicity} 
& \multicolumn{6}{c}{\large Gender} \\
\cline{4-15}
& & & \multicolumn{2}{c|}{\large MaMA $\uparrow$} & \multicolumn{2}{c|}{\large MiMA $\uparrow$} & \multicolumn{2}{c|}{\large  HomE $\downarrow$} 
& \multicolumn{2}{c|}{\large MaMA $\uparrow$} & \multicolumn{2}{c|}{\large MiMA $\uparrow$} & \multicolumn{2}{c}{\large HomE $\downarrow$} \\
\cline{4-15}
& & & \rotatebox{60}{BFW} &  \rotatebox{60}{BUPT} & 
\rotatebox{60}{BFW} & \rotatebox{60}{BUPT} & 
\rotatebox{60}{BFW} & \rotatebox{60}{BUPT} & 

\rotatebox{60}{BFW} &  \rotatebox{60}{BUPT} & 
\rotatebox{60}{BFW} &  \rotatebox{60}{BUPT} & 
\rotatebox{60}{BFW} & \rotatebox{60}{BUPT} \\
\hline
\rowcolor{red!15}
FairFace~\cite{Karkkainen2021FairFace}   & 21.3 & - & 0.90 & 0.88 & 0.90 & 0.88 & 0.19 & 0.22 & 0.96 & \underline{\textbf{\textit{0.98}}} &  0.96 & 0.97 & 0.19 & 0.22 \\
\rowcolor{red!15} FairFace (reprod)&     21.3 & - & 0.80 & 0.79 & 0.80 &  0.79 & 0.38 & 0.41  &  0.80  & 0.79 &  0.80 & 0.79 & 0.25   & 0.19 \\
\rowcolor{red!15} BUPT (reprod)&       21.3 & - & 0.80 & 0.90 & 0.80 & 0.89  & 0.35 & 0.18  & 0.94 & 0.97  &  0.95 &  0.96 &0.19   & 0.12 \\
\rowcolor{red!15} DeepFace\cite{serengil2024benchmark}& 134 & - & 0.84 & 0.80 & 0.84 &  0.80 & 0.29 & 0.36 &  0.90 & 0.94 & 0.90 &  0.95 & 0.31  & 0.16\\
\rowcolor{green!15} RN34+RBF     & 21.3 & 1 & 0.93 & 0.91 & 0.93 & 0.90 & 0.14 & 0.17  & 0.96 & \underline{\textbf{\textit{0.98}}} & 0.96 & 0.97 & 0.16 & 0.11 \\
\rowcolor{green!15} IR18+RBF     & 24.0 & 1 & 0.94 & 0.92 & 0.94 & 0.93 & 0.13 & 0.14  & 0.96 & \underline{\textbf{\textit{0.98}}} & \underline{\textbf{\textit{0.97}}} & \textit{\textbf{0.98}} & 0.14 & \textit{0.08} \\
\rowcolor{green!15} RN50+RBF     & 24.6 & 1 &  \textit{0.95} & \textit{0.94} & \textit{0.95} & \textit{0.94} & \textit{0.09} & 0.11  & \underline{\textbf{\textit{0.97}}} & \underline{\textbf{\textit{0.98}}} & \underline{\textbf{\textit{0.97}}}& \underline{\textbf{\textit{0.98}}} & \underline{\textbf{\textit{0.12}}} & \textit{0.08} \\
\rowcolor{green!15} ViT-T+RBF    & 19.1 & 1 & \textit{0.95} & 0.93 & \textit{0.95} & \textit{0.94} & 0.10 & \textit{0.10} & 0.96 & \underline{\textbf{\textit{0.98}}} & \underline{\textbf{\textit{0.97}}} & \underline{\textbf{\textit{0.98}}} & 0.14 & \textit{0.08} \\
\rowcolor{blue!15} RN101+RBF     & 43.5 & 1 & \textbf{\underline{0.96}} & 0.94 & 0.96 & 0.94 & 0.09 & 0.11 
& \textbf{\underline{0.97}} & \textbf{\underline{0.98}} & \textbf{\underline{0.97}} & \textbf{\underline{0.98}} & 0.13 & 0.08  \\
\rowcolor{blue!15}  IR50+LIN     & 43.5 & 1 & 0.91 & 0.89 & 0.91 & 0.90 & 0.19 & 0.20 & 0.95 & 0.97 & 0.95 &  0.97 & 0.19  &  0.13\\
\rowcolor{blue!15} IR50+RBF      & 43.6 & 1 & \textbf{\underline{0.96}} & \textbf{\underline{0.95}} & \textbf{\underline{0.96}} & \textbf{\underline{0.95}} & \underline{\textbf{0.08}} & \underline{\textbf{0.09}} & 0.96 & \textbf{\underline{0.98}} & \textbf{\underline{0.97}} & \textbf{\underline{0.98}} & \underline{\textbf{0.12}} & \underline{\textbf{0.07}} \\
\rowcolor{blue!15} ViT-S+RBF     & 76.0 & 1 & \underline{\textbf{0.96}} & 0.94 & \textbf{\underline{0.96}} & \textbf{\underline{0.95}} & 0.11 & \underline{\textbf{0.09}} 
& \textbf{\underline{0.97}} & \textbf{\underline{0.98}} & \textbf{\underline{0.97}} & \textbf{\underline{0.98}} & 0.13 &  \underline{\textbf{0.07}} \\
\rowcolor{orange!15} IR18+RBF       & 24.0 & 40  & 0.93 & 0.92 & 0.94 & 0.93 & 0.13 & 0.14  & \textbf{0.97} & \textbf{0.98} & \textbf{0.97} & \textbf{0.98} &  0.13 & 0.08 \\
\rowcolor{orange!15}    IR50+RBF    & 43.6 & 40 & 0.91 & 0.92 & 0.91 & 0.92 & 0.18 & 0.16 & 0.96 &  \textbf{0.98} & \textbf{0.97} & \textbf{0.98} &  0.13 & 0.08 \\

\end{tabular}
}
\caption{Results with all robustness metrics. Column 3 (\#imgs) is the number of images used for pretraining (in millions). See~\autoref{tab_results} for detail son models. The best results for compact models comparable with FairFace~\cite{Karkkainen2021FairFace} (red and green rows) are in \textit{italics}. The best results obtained when including larger pretrained architectures (red, green, and blue rows) are \underline{underlined}. The best global results in \textbf{bold}. Higher figures are better for accuracy, while lower figures are better for fairness and robustness.}\label{tab:supp_all_robustness}
\end{table*}



\section{Details on Theoretical Aspects}\label{sec:sup_theory_proof}

\subsection{Proofs of Propositions}
\paragraph{Notations}

We consider a face analysis system (FAS) trained on an images dataset, each of these last being annotated by a class label (either binary on pairs of images for face verification or multicategorial for each image for face recognition) and a set of $N_a$ sentitive attributes, each being categorial. The cartesian product of all possible classes of $N_a$ attributes leads to classify the images into $K$ groups\footnote{for example, it there are 4 ethnicities, 2 genders and 4 classes of age, then $K=4*2*4=32$}.

Let $G\in\{1,\dots,K\}$ be the true latent group of an identity, $\widehat G\in\{1,\dots,K\}$ the estimated label of the corresponding sensitive attribute, and $Y\in\{0,1\}$ a binary performance indicator associated to an identity for the considered FAS. 

We define the prior of the sensitive attribute $a$ as
\begin{equation}
    \pi_a := \Pr(G=a)
\end{equation}
and the probability of success of the FAS given the \textit{true} said sensitive attribute as
\begin{equation}
p_a := \Pr(Y=1\mid G=a)
\end{equation}
This last is actually estimated automatically with some attribute classifiers that have a confusion matrix $C=(c_{ab})_{a,b=1}^K$ with
\begin{equation}
    c_{ab} := \Pr(\widehat G=b \mid G=a)
\end{equation}
such that each row of $C$ sums to 1. We then define:
\begin{align}
    \tau_b & := \Pr(\widehat G=b) = \sum_{a=1}^K \pi_a c_{ab} \label{eq:def_tau_b}\\
    m_g & := \Pr(Y=1\mid \widehat G = g)
\end{align}
For convenience of notations we set:
\begin{align}
    p&=(p_1,\dots,p_K)^\top\\
    \pi&=(\pi_1,\dots,\pi_K)^\top \\
    \tau&=(\tau_1,\dots,\tau_K)^\top=C^\top\pi
\end{align}
and $\odot$ is the elementwise product. 
We consider an i.i.d. sample of $I$ identities; for identity $i$ we observe $(\widehat G_i,Y_i)$ but not $G_i$. For each observed label $g$ we define the empirical plug-in estimator:
\begin{equation}\label{eq:empirical_pugin_estim}
\widehat m_g = \frac{\frac{1}{I}\sum_{i=1}^I \mathbf{1}\{\widehat G_i=g\} Y_i}{\frac{1}{I}\sum_{i=1}^I \mathbf{1}\{\widehat G_i=g\}}
\end{equation}

We note  $\widehat m=(\widehat m_1,\dots,\widehat m_K)^\top$ assuming the denominator is nonzero for the $K$ groups considered, and $\widehat \tau$ the empirical estimator of $\tau$.


\paragraph{\autoref{theorem:bias} (formal)}

If one assumes that $(Y \perp\!\!\!\perp \widehat G)\mid G$ (\autoref{eq:assume_conditinal_inped}) and $C^\top$ (thus $C$) is invertible, then: 
\begin{itemize}
    \item the empirical plug-in estimator $\widehat m_g$ converges in probability toward $m_g$ as $I\to\infty$ and:
\end{itemize}
\begin{equation}
    m_g = \frac{\sum_{a=1}^K \pi_a c_{a g} p_a}{\sum_{a=1}^K \pi_a c_{a g}} = \frac{1}{\tau_g}\sum_{a=1}^K\pi_ac_{ag}p_a
\end{equation}
\begin{itemize}
    \item the bias of the plug-in limit relative to the true group rate $p_g$ satisfies:
\end{itemize}
\begin{equation}\label{eq:propbias_bias}
\mathrm{Bias}_g := m_g - p_g
= \frac{\sum_{a\ne g} \pi_a c_{a g} (p_a-p_g)}{\sum_{a=1}^K \pi_a c_{a g}}
\end{equation}
and therefore if $c_{ag}\to 0$ for every $a\ne g$ and $c_{g g}\to 1$ then $\mathrm{Bias}_g\to 0$. 
Moreover, for $\Delta_{\max,g} := \max_{a\ne g} |p_a-p_g|$,
\begin{equation}\label{eq:propbias_bound}
    \big|\mathrm{Bias}_g\big|
\le \Delta_{\max,g}\cdot \frac{\sum_{a\ne g}\pi_a c_{a g}}{\pi_g c_{g g}}
\end{equation}

\begin{proof}
For a fixed $g$, by the weak law of large numbers we have the convergence in probability of the numerator and denominatior of \autoref{eq:empirical_pugin_estim}:
\begin{equation}
    \frac{1}{I}\sum_{i=1}^I \mathbf{1}\{\widehat G_i=g\} Y_i \xrightarrow{p} \mathbb{E}[\mathbf{1}\{\widehat G=g\} Y] = \Pr(\widehat G=g,\, Y=1)
\end{equation}
and
\begin{equation}
    \frac{1}{I}\sum_{i=1}^I \mathbf{1}\{\widehat G_i=g\} \xrightarrow{p} \Pr(\widehat G=g)=\tau_g
\end{equation}
provided $\tau_g>0$. Therefore the ratio $\widehat m_g$ converges in probability to
\begin{equation}
    \frac{\Pr(\widehat G=g,\,Y=1)}{\Pr(\widehat G=g)} = \Pr(Y=1\mid \widehat G=g)
\end{equation}
that is $m_g$ by definition. By the law of total probability and \autoref{eq:assume_conditinal_inped} we have:
\begin{align}
    \Pr(\widehat G=g,\,Y=1) = \sum_{a=1}^K \Pr(\widehat G=g,\,Y=1,\,G=a) \nonumber \\
    = \sum_{a=1}^K \Pr(\widehat G=g\mid G=a)\Pr(Y=1\mid G=a)\Pr(G=a) \nonumber \\
    = \sum_{a=1}^K c_{a g}\,p_a\,\pi_a 
\end{align}
Using \autoref{eq:def_tau_b}:
\begin{equation}
    m_g = \frac{\sum_{a=1}^K \pi_a c_{a g} p_a}{\sum_{a=1}^K \pi_a c_{a g}}
\end{equation}
that shows the first point of the proposition.

By definition,
\begin{align}
    \mathrm{Bias}_g & = m_g - p_g \\
& = \frac{\sum_{a=1}^K \pi_a c_{a g} p_a}{\sum_{a=1}^K\pi_a c_{a g}} - p_g \\
& = \frac{\sum_{a=1}^K \pi_a c_{a g} (p_a-p_g)}{\sum_{a=1}^K\pi_a c_{a g}} \\
\end{align}
That is the same as \autoref{eq:propbias_bias} since the term for $a=g$ vanished in the sum of the numerator. Applying the triangle inequality to the absolute value, we have:
\begin{align}
    |\mathrm{Bias}_g| & = \left| \frac{\sum_{a=1}^K \pi_a c_{a g} (p_a-p_g)}{\sum_{a=1}^K\pi_a c_{a g}}\right| \\
    & \le \frac{\sum_{a=1}^K \pi_a c_{a g} |p_a-p_g|}{\sum_{a=1}^K\pi_a c_{a g}} \\
    & \le \Delta_{\max,g}\cdot \frac{\sum_{a\ne g} \pi_a c_{a g}}{\sum_{a=1}^K\pi_a c_{a g}}
\end{align}
Since $\sum_{a=1}^K\pi_a c_{a g}\ge \pi_g c_{g g}$, we finally have the bound:
\begin{equation}
    |\mathrm{Bias}_g|
\le \Delta_{\max,g}\cdot \frac{\sum_{a\ne g} \pi_a c_{a g}}{\pi_g c_{g g}}
\end{equation}
That is the same as \autoref{eq:propbias_bound}. Hence, if the confusion matrix tends towards identity then $c_{ag}\to 0$ for every $a\ne g$ and $c_{gg}\to 1$, thus the numerator $\sum_{a\ne g}\pi_a c_{a g}\to 0$ and denominator $\sum_a\pi_a c_{a g}\to \pi_g$, so $\mathrm{Bias}_g\to 0$.
\end{proof}

\paragraph{\autoref{theorem:covariance} (formal)}
If $C^\top$ is invertible then the estimator
\begin{equation}
  \widehat p^{\mathrm{(corr)}} \;=\; \frac{(C^\top)^{-1} (\widehat\tau\odot\widehat m)}{\pi}
\end{equation}
is asymptotically unbiased for $p$. For operator norms $\|\cdot\|_{\mathrm{op}}$, its covariance matrix satisfies:


\begin{equation}
    \|\operatorname{Cov}\big((C^\top)^{-1}(\widehat\tau\odot\widehat m)\big)\|_{\mathrm{op}}
\le \|(C^\top)^{-1}\|_{\mathrm{op}}^2 \cdot \|\operatorname{Cov}(\widehat\tau\odot\widehat m)\|_{\mathrm{op}}.
\end{equation}
Thus the factor $\|(C^\top)^{-1}\|_{\mathrm{op}}^2$ controls the worst-case multiplicative increase in variance; when $C\to I$ the factor tends to $1$, while ill-conditioned $C$ make $\|(C^\top)^{-1}\|_{\mathrm{op}}$ large and amplify variance.

\begin{proof}

Let us consider the population-observed numerator vector $n\in\mathbb{R}^K$ with $n_g := \Pr(\widehat G=g,\,Y=1) = \sum_{a=1}^K \pi_a c_{a g} p_a$. We have $m_g=n_g/\tau$ and hence $n=\tau\odot m$. Thus:
\begin{align}
    n &= C^\top (\pi\odot p) \\
    \tau\odot m &= C^\top (\pi\odot p)
\end{align}
Thus if $C^\top$ is invertible:
\begin{align}
    \pi\odot p &= (C^\top)^{-1} (\tau\odot m) \\
p &= \frac{(C^\top)^{-1} (\tau\odot m)}{\pi} 
\end{align}
where the second line is an elementwise division.  This is an algebraic identity at the population level. Accordingly, an estimator that replaces $\tau$ and $m$ by consistent estimators $\widehat\tau,\widehat m$ and plugs in $C$ converges to $p$. In particular, if $C$ is known exactly and $\widehat\tau\odot\widehat m$ is (asymptotically) unbiased and consistent for $\tau\odot m$, then $(C^\top)^{-1}(\widehat\tau\odot\widehat m)$ is (asymptotically) unbiased and consistent for $\pi\odot p$, and dividing elementwise by known $\pi$ yields an unbiased consistent estimator of $p$.

 Let $\widehat n := \widehat\tau\odot\widehat m$ be the empirical estimator of $n$. For a large number of identities, $\widehat n$ is approximately unbiased with some covariance matrix $\Sigma_{\widehat n}=\operatorname{Cov}(\widehat n)$. For the the (asymptotic) covariance of the corrected estimator we have:
 \begin{equation}
     \operatorname{Cov}\big((C^\top)^{-1}\widehat n\big) = (C^\top)^{-1}\;\Sigma_{\widehat n}\; (C^\top)^{-T}.
 \end{equation}
Both sides are square matrix on Hilbert spaces thus, taking the operator norms, we have:
\begin{equation}
\big\|\operatorname{Cov}\big((C^\top)^{-1}\widehat n\big)\big\|_{\mathrm{op}}
\le \|(C^\top)^{-1}\|_{\mathrm{op}}^2 \cdot \|\Sigma_{\widehat n}\|_{\mathrm{op}}.
\end{equation}
Thus the matrix norm $\|(C^\top)^{-1}\|_{\mathrm{op}}^2$ controls the worst-case multiplicative inflation of covariance when inverting $C^\top$.If $C$ tends toward identity, (i.e. the estimation of the attributes becomes perfect), then $\|(C^\top)^{-1}\|_{\mathrm{op}}\to 1$ and  there is thus no variance inflation. Conversely, if $C$ has very small singular values (\ie $C$ is ill-conditioned), then $\|(C^\top)^{-1}\|_{\mathrm{op}}$ is large and the variance of the corrected estimator can be dramatically larger than the variance of $\widehat n$.

\end{proof}

\subsection{Binary Indicator and Fairness Metrics}\label{sec:link_Yi_metrics}
There exist several metrics to estimate the fairness of face analysis systems (FAS), as presented in~\autoref{sec:supp_metrics_fairness}. For a group $g$, the False Match Rate (FMR) is ``the rate at which a biometric process mismatches biometric signals from two distinct individuals as coming from the same individual''~\cite{schuckers2010false_match_rate}. It is usually used for face verification, that is from a pair of face images, although some works consider this metrics for face recognition. For a pair of identities, it is defined as:
\begin{equation}
       \text{FMR}_g = \Pr(\text{FAS accepts} \mid \text{pair is impostor},\, G=g).
\end{equation}
 
Common fairness metrics aims at reflecting the Demographic Parity, which ensures that the outcome of the FAS is the same across all groups defined by a sensitive attribute. Hence, the fairness can be estimated with the Demographic Parity Difference (DPD) and Demographic Parity Ratio (DPR) defined as
\begin{align}
       \text{DPD}_{\text{FMR}} & = \max_{g,g'} |\text{FMR}_g - \text{FMR}_{g'}| \forall g,g'\in G\\
       \text{DPR}_{\text{FMR}} & = \frac{\min_g FMR_g}{\max_g FMR_g}
\end{align}
A difference close to 0 or a ratio close to 1 indicate a decision system fair with respect to the sensitive attribute accross groups~\cite{castelnovo2022_clarify_fairness}.

Another common fairness metric is the Degree of Bias (DoB), defined as the accuracy across different subgroups, which is higher when the performance varies a lot w.r.t each subgroup. Hence it can be expressed from the FMR as:
\begin{equation}
       \text{DoB} = \sqrt{\frac{1}{K} \sum_{g=1}^K \left(1 - \frac{\text{FMR}_g}{\overline{\text{FMR}}}\right)^2 },
\end{equation}
 where $\overline{\text{FMR}}$ is the average FMR over groups.

The generic performance indicator $Y_i \in \{0,1\}$ we consider in our paper can be specializer to express these metrics. If one evaluates the impostor trials such that $Y_i = \mathbbm{1}\{\text{FRS accepts impostor trial}\}$,  we have $\Pr(Y=1 \mid G=g) = \text{FMR}_g$. If one  rather evaluates the genuine trials then $Y_i = 1\{\text{FRS rejects genuine trial}\}$, thus  we have the False Non-Match Rate (FNMR)\footnote{i.e 1 - True Match Rate (TMR)}: $\Pr(Y=1 \mid G=g) = \text{FNMR}_g$. 
Thus the group-specific expectation $p_g := \Pr(Y=1\mid G=g)$ is exactly the per-group error rate. The fairness metrics can thus be expressed accordint to the vector $p=(p_1,\dots,p_K)$, \eg:
 \begin{align}
     \text{DPD}_{\text{FMR}} & = \max_{g,g'} |p_g - p_{g'}| \\
     \text{DoB} & = \sqrt{\frac{1}{K}\sum_{g=1}^K \left(1 - \frac{p_g}{\overline{p}}\right)^2}
 \end{align}


\end{document}